\documentclass[letterpaper, 10 pt, conference]{ieeeconf}
\IEEEoverridecommandlockouts
\usepackage[english]{babel}
\usepackage[pdftex]{graphicx}
\usepackage{adjustbox}
\graphicspath{{./figures/}}
\DeclareGraphicsExtensions{.pdf,.jpg,.jpeg,.png}
\usepackage[cmex10]{amsmath}
\usepackage{bm}
\interdisplaylinepenalty=2500{}
\usepackage{physics}
\usepackage{booktabs,multirow}
\usepackage[caption=false,font=footnotesize]{subfig}
\def\enquote#1{``{#1}''}
\usepackage{tikz,pgfplots}
\pgfplotsset{compat=1.3}
\usepackage{url}
\newcommand{\fref}[1]{Fig.~\ref{#1}}
\newcommand{\sfref}[2]{\fref{#1}\,\subref{#2}}
\newcommand{\tref}[1]{Table~\ref{#1}}
\newcommand{\sref}[1]{Section~\ref{#1}}
\newcommand{\mdof}{\text{\ac{dof}}}

\usepackage[acronym,shortcuts]{glossaries}
\glsdisablehyper
\newacronym{cip}{CIP}{Constraint Integer Programming}
\newacronym{dfs}{DFS}{Depth-First Search}
\newacronym[longplural=Degrees of Freedom]{dof}{DoF}{Degree of Freedom}
\newacronym{ga}{GA}{Genetic Algorithm}
\newacronym{glkh}{GLKH}{Generalized Lin-Kernighan-Helsgaun}
\newacronym{gtsp}{GTSP}{Generalized Traveling Salesman Problem}
\newacronym{ik}{IK}{Inverse Kinematics}
\newacronym{lkh}{LKH}{Lin-Kernighan-Helsgaun}
\newacronym{ros}{ROS}{Robot Operating System}
\newacronym{rrt}{RRT}{Rapidly-Exploring Random Tree}
\newacronym{birrt}{BiRRT}{Bidirectional Rapidly-Exploring Random Tree}
\newacronym{rnn}{RNN}{Recursive Nearest-Neighbor}
\newacronym{rtsp}{RTSP}{Robotic Task Sequencing Problem}
\newacronym{scip}{SCIP}{Solving Constraint Integer Programs}
\newacronym{tsp}{TSP}{Traveling Salesman Problem}
\glsunset{dof}

\title{\LARGE
  \textbf{RoboTSP -- A Fast Solution to the Robotic Task Sequencing Problem}}

\author{Francisco Su\'{a}rez-Ruiz, Teguh Santoso Lembono and Quang-Cuong Pham%
  \thanks{The authors are with the School of Mechanical and Aerospace
          Engineering, Nanyang Technological University, Singapore.}}

\begin{document}
\maketitle
\thispagestyle{empty}
\pagestyle{empty}

\begin{abstract}
  In many industrial robotics applications, such as spot-welding, spray-painting
  or drilling, the robot is required to visit successively multiple targets. The
  robot travel time among the targets is a significant component of the overall
  execution time. This travel time is in turn greatly affected by the order of
  visit of the targets, and by the robot configurations used to reach each
  target. Therefore, it is crucial to optimize these two elements, a problem
  known in the literature as the \ac{rtsp}. Our contribution in this paper is
  two-fold. First, we propose a fast, near-optimal, algorithm to solve
  \ac{rtsp}. The key to our approach is to exploit the classical distinction
  between task space and configuration space, which, surprisingly, has been so
  far overlooked in the \ac{rtsp} literature. Second, we provide an open-source
  implementation of the above algorithm, which has been carefully benchmarked to
  yield an efficient, ready-to-use, software solution. We discuss the
  relationship between \ac{rtsp} and other \ac{tsp} variants, such as the
  \ac{gtsp}, and show experimentally that our method finds motion sequences of
  the same quality but using several orders of magnitude less computation time
  than existing approaches.
\end{abstract}


\section{Introduction}
\label{sec:introduction}

\begin{figure}[t]
  \centering
  \vspace*{2mm}
  \subfloat[]{\includegraphics[height=35mm]{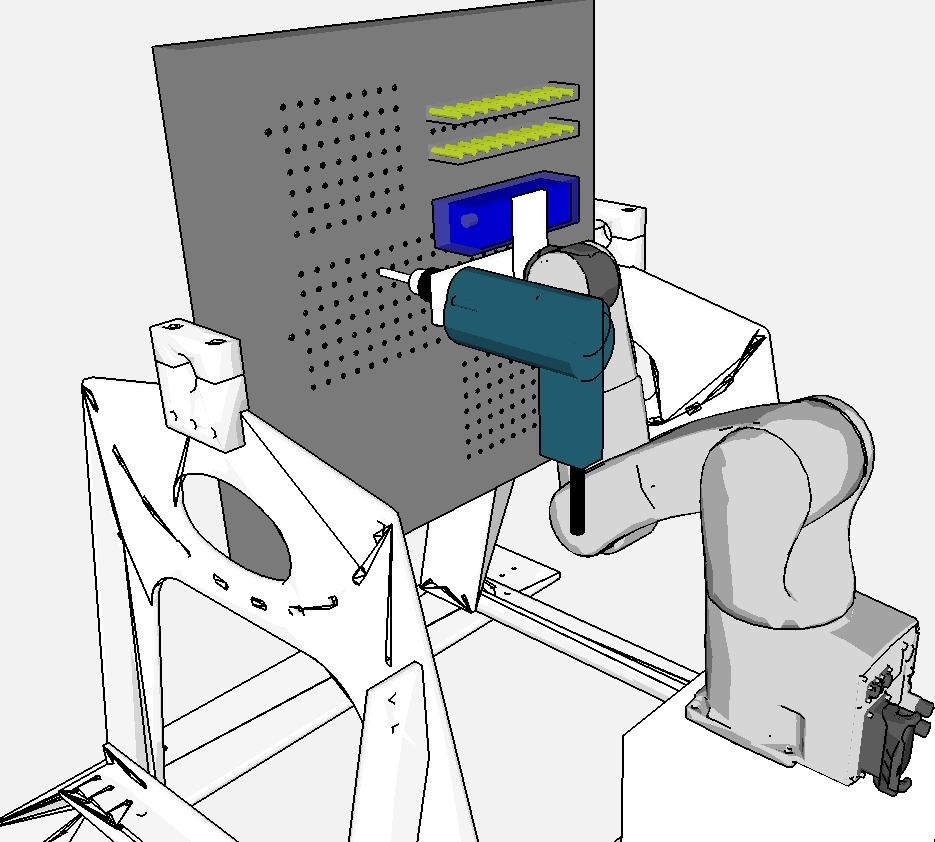}}\qquad
  \subfloat[]{\includegraphics[height=35mm]{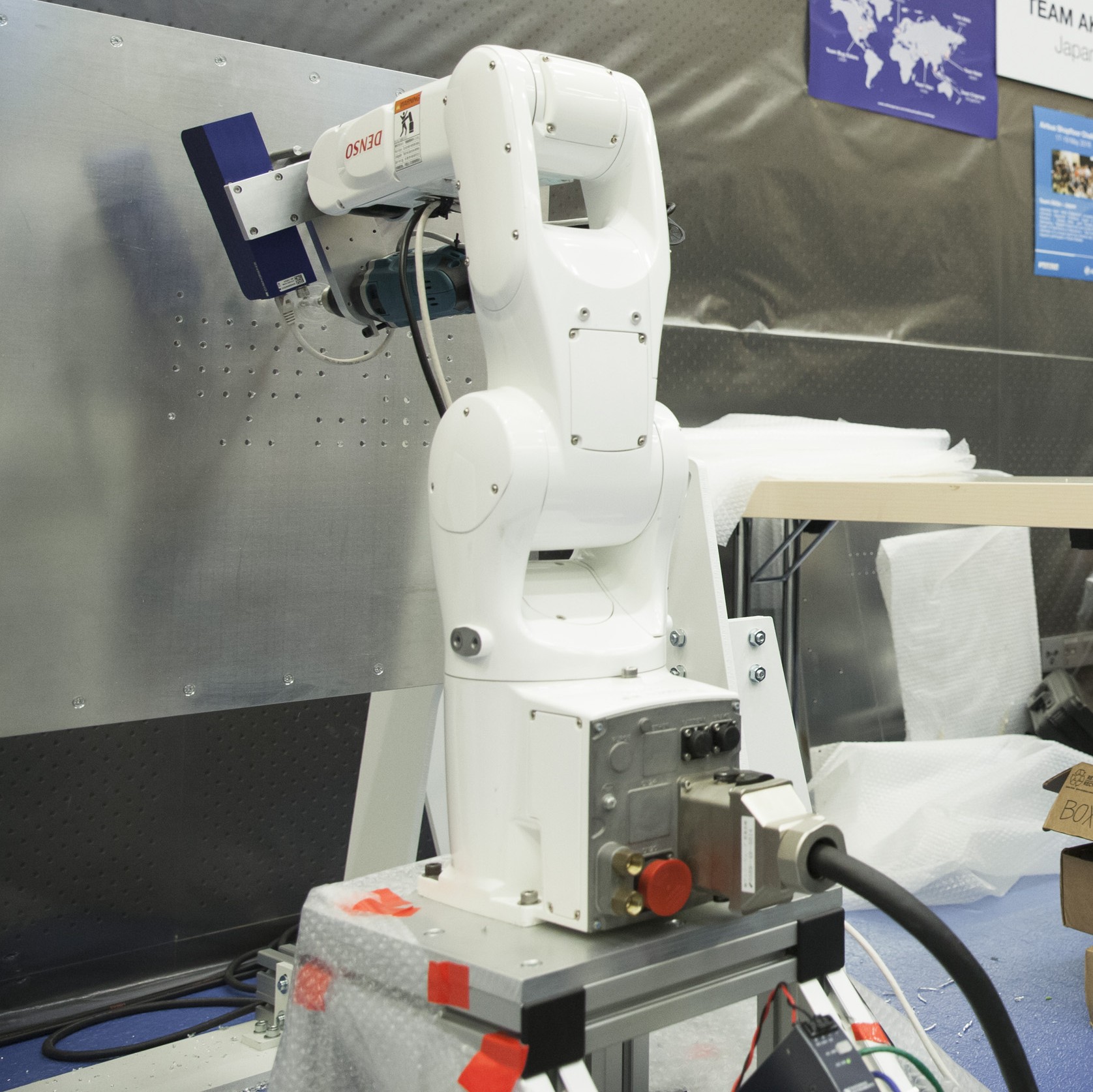}}
  \caption{The Airbus Shopfloor Challenge held during ICRA 2016. The task,
  mimicking the actual drilling process in aircraft manufacturing, consisted in
  drilling as many holes as possible, in one hour, from a given pattern of $245$
  hole positions. Since the position of the panel is unknown at the beginning of
  the round, the planning time is included within the one hour limit.
  a) The simulation environment. b) Our setup using a DENSO
  robot at the live Challenge, where our team won the second prize.}
  \label{fig:setup}
\end{figure}

In many industrial robotics applications, such as spot-welding, spray-painting
or drilling, the robot is required to visit successively multiple targets.
Consider for instance the drilling task depicted in~\fref{fig:setup}, which was
proposed at the Airbus Shopfloor Challenge held during ICRA 2016 in Stockholm,
Sweden. The task, mimicking the actual drilling process in aircraft
manufacturing, consisted in drilling as many holes as possible in one hour,
from a given pattern of $245$ hole positions. The robot travel time between
the holes is a significant component of the overall execution time. This travel time
is in turn greatly affected by the order of visit of the holes, and by the robot
configurations used to reach each hole. Therefore, it is crucial to optimize
these two elements, a problem known in the literature as the \ac{rtsp},
see~\cite{Alatartsev2015} for a recent review. Finally, note that, since the
position of the panel is unknown at the beginning of the round, the planning
time is included within the one hour limit. Thus, in this Challenge as in many
practical applications, there is a need for an algorithm that can find
near-optimal plans within minutes, not hours.

\begin{figure}[t]
  \centering
  \vspace*{2mm}
  \subfloat[]{\includegraphics[width=0.45\linewidth]{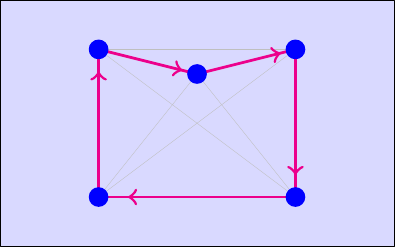}
  \label{subfig:tsp}}\;
  \subfloat[]{\includegraphics[width=0.45\linewidth]{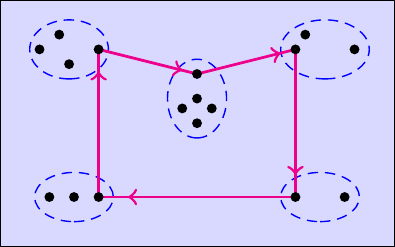}
  \label{subfig:gtsp}}\\
  \subfloat[]{\includegraphics[width=0.8\linewidth]{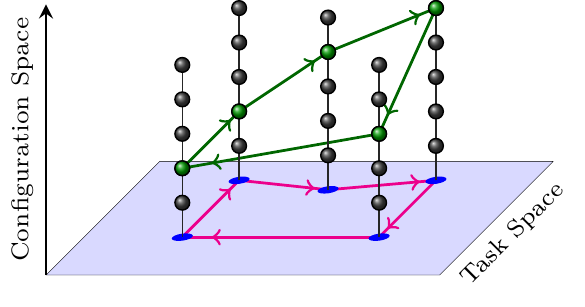}
  \label{subfig:rtsp}}
  \caption{\protect\subref{subfig:tsp} The \acl{tsp}.
  \protect\subref{subfig:gtsp} The \acl{gtsp}. \protect\subref{subfig:rtsp}
  Decomposition of the \acl{rtsp} into task-space and configuration-space.}
  \label{fig:layouts}
\end{figure}

\ac{rtsp} is closely related to \ac{tsp}, a classical problem in Computer
Science. In \ac{tsp}, a set of locations is given on a map, and one is
interested in finding the order to visit all the locations while minimizing the
total travel distance, see~\sfref{fig:layouts}{subfig:tsp}. The key difference
between \ac{rtsp} and \ac{tsp} is that, in \ac{rtsp}, each of the targets (e.g.
a hole position) may be reached by \emph{multiple robot configurations}, also
known as \emph{\ac{ik} solutions}.

One simple work-around may consist in assigning a fixed robot configuration for
each target: \ac{rtsp} then becomes a classical \ac{tsp} among the assigned
robot configurations. Such a work-around is however sub-optimal. Another
approach consists in formulating \ac{rtsp} as an instantiation of \ac{gtsp}:
in \ac{gtsp}, the locations are split into bins, and one is required to visit
exactly one location per bin,
see~\sfref{fig:layouts}{subfig:gtsp}. Here, each bin will contain
the different robot configurations corresponding to the same target. While there
have been many works devoted to \ac{gtsp} and some efficient solutions
exist, the sheer size of real-world \ac{rtsp} instances\,\footnote{For instance,
in the set-up of the Airbus Shopfloor Challenge, using a discretization of
$\frac{\pi}{2}$ radians for the free-\ac{dof}, one obtains $3,779$ different
configurations, grouped into $245$ bins, which cannot be solved in practical
times by any existing \ac{gtsp} solver. See also \sref{sub:other_methods} for a
detailed comparison.} make this approach inapplicable in practice.
\sref{sec:related} provides a more detailed discussion on existing approaches to
solve \ac{rtsp}.

Our contribution in this paper is two-fold. First, we propose a fast,
near-optimal, algorithm to solve \ac{rtsp}. The key to our approach is to
exploit the classical \emph{distinction between task space and configuration
space}, which, surprisingly, has been so far overlooked in the \ac{rtsp}
literature. Specifically, we propose a three-step algorithm
[see~\sfref{fig:layouts}{subfig:rtsp} for illustration]: \begin{enumerate}

  \item Find a (near-)optimal visit order of the targets in a \emph{task-space
  metric} (e.g. Euclidean distance between the hole positions), using classical
  \ac{tsp} algorithms;

  \item Given the order found in Step~1, find for each target the optimal robot
  configuration, so that the total path length through the configurations is
  minimized in a \emph{configuration-space metric} (e.g. Euclidean distance
  between the robot configurations -- collisions are ignored at this stage),
  using a graph shortest path search algorithm;

  \item Compute the final collision-free configuration-space trajectories by
  running classical motion planning algorithms (e.g. \ac{rrt}) through the robot
  configurations found in Step~2 and in the order given by Step~1.

\end{enumerate}

Our second contribution is to provide an open-source implementation\,\footnote{The
open-source implementation of the proposed method is accessible at
\url{www.github.com/crigroup/robotsp}} of the above algorithm. In particular, we
carefully benchmark different key components of the algorithm (underlying
task-space \ac{tsp} solver, configuration-space metric, discretization step-size
for the free-\ac{dof}), so as to come up with an efficient, ready-to-use,
software solution.

The remainder of the paper is organized as follows. In \sref{sec:related}, we
discuss existing approaches to \ac{rtsp}. In \sref{sec:method}, we introduce and
describe in detail the proposed method. In \sref{sec:experiments}, we present
the experimental results, showing in particular that our method finds motion
sequences of the same quality but using several orders of magnitude less
computation time than existing approaches. Finally, in \sref{sec:conclusions},
we conclude with a few remarks.

\section{Related works}
\label{sec:related}

A recent survey on strategies to solve \ac{rtsp} can be found in
\cite{Alatartsev2015}. We recall some of the main results below.

In~\cite{Abdel-Malek1990}, the authors plan the motions for a 3-\ac{dof} robot
to visit $6$ targets, each of which can be reached with two different robot
configurations. For this, they formulate a \ac{gtsp} with $6$ bins and $2$
configurations per bin. They then convert the \ac{gtsp} into a \ac{tsp}, which
can be solved efficiently.

In \cite{Edan1991}, the authors consider a fruit picking task, which has up to
$250$ targets, but with only one robot configuration per target. The problem can
then be directly formulated as a \ac{tsp}, which the authors solved using the
Nearest-Neighbor heuristic.

In \cite{Saha2003}, the authors consider a \ac{rtsp} with one configuration per
target, which can then be formulated as a regular \ac{tsp}. The emphasis here is
on collision-free trajectories, which are difficult to find when the environment
includes obstacles. First, the authors approximate the travel cost between
configurations as the Euclidean distance in the configuration space. Then, a
minimum spanning tree is computed using Prim's algorithm to find a near-optimal
tour under the approximated cost. Next, collision-free paths are calculated
given this near-optimal tour. The collision-free paths yield an updated travel
cost, which is then used to iteratively refine the tour. This idea of computing
first a good tour with a simple metric (without considering collisions) before
computing collision-free trajectories is reused in our present work (but without
the iterative refinement step).

An extension of this work is proposed in \cite{Saha2006}. Here the authors
consider multiple robot configurations per target. Instead of a spanning tree,
they compute a near-optimal \emph{group spanning tree} \cite{Chekuri2006}. On a
task involving $50$ targets with $5$ configurations per target, a near-optimal
solution could be found in $9,600$\,s.

Following a different approach, the authors of \cite{Zacharia2005} propose to
use \acp{ga} to solve \ac{gtsp}. The optimization criteria is the task cycle
time. On a task with 3-\ac{dof} and 6-\ac{dof} robots involving $50$ targets, a
near-optimal solution is found in $1,800$\,s. The quality of the solution
depends on several control parameters (related to the \ac{ga}) and the number of
iterations. This approach has been further extended to include collision-free
path planning for 2D and 3D environments \cite{Xidias2010,Zacharia2013}.

Yet another approach consists in formulating \ac{rtsp} as a multi-objective
constraint optimization problem. In \cite{Kolakowska2014}, a robotic spray
painting task is considered. The optimization criteria is set to minimize the
task planning and execution time while maximizing the painting quality. Three
constraints were defined: process, resources and quality constraints. The
multi-objective problem then is solved using the \ac{dfs} algorithm. On a task
involving $8$ targets, with $4$ configurations per target, an optimal solution
is found in $10,000$\,s.

The main limitation of all the works discussed above is the large computation
time they require. In particular, none of these works could have been applied to
the setting of the Airbus Shopfloor Challenge presented in
\sref{sec:introduction}, which involves $245$ targets, with tens of configurations
per target, and which has to be solved within a few minutes -- since the planning
time is counted in the one hour limit of the challenge.


\section{RoboTSP algorithm}
\label{sec:method}

\subsection{Setting}

Consider $n$ targets in the task-space. A tour in the task-space that visits
each target exactly once is called a \emph{task-space tour}\,\footnote{Strictly
speaking, a tour requires to return to the first target, so we are making a
slight abuse of vocabulary here.}. We first compute \ac{ik} solutions for each
target -- using a suitable discretization for the free-\acp{dof} if necessary. A
tour in the configuration-space that starts from the robot \enquote{home}
configuration, visits, for each target, exactly one \ac{ik} solution associated
with that target, and returns to the \enquote{home} configuration is called a
\emph{configuration-space tour}. Our objective is to find the fastest
\emph{collision-free} configuration-space tour subject to the robot constraints
(e.g. velocity and acceleration bounds).

Let $m_i$ be the number of \ac{ik} solutions found for target~$i$.  If we do not
take into account obstacles, there are $\qty(n-1)!\qty(\prod_{i=1}^{n} m_{i})$
possible configuration-space tours (with straight paths) for this task. One
cannot therefore expect to find the optimal sequence by brute force in practical
times.

\subsection{Proposed algorithm}

As presented in \sref{sec:introduction}, our method consists in:

\begin{enumerate}

  \item Finding a (near-)optimal task-space tour in a \emph{task-space metric};

  \item Given the order found in Step~1, finding, for each target, the optimal
  robot configuration, so that the corresponding configuration-space tour has
  minimal length in a \emph{configuration-space metric} -- collisions are
  ignored at this stage;

  \item Computing fast collision-free \textit{configuration-space} trajectories
  by running classical motion planning algorithms (e.g. \ac{rrt}-Connect with
  post-processing~\cite{Kuffner2000a,Pham2015}) through the configurations found
  in Step~2 and in the order given by Step~1.

\end{enumerate}

Implementation details and benchmarking results for Steps 1 and 3 are given in \sref{sec:experiments}.

\begin{figure}[t]
  \centering
  \vspace*{2mm}
  \includegraphics[width=0.95\linewidth]{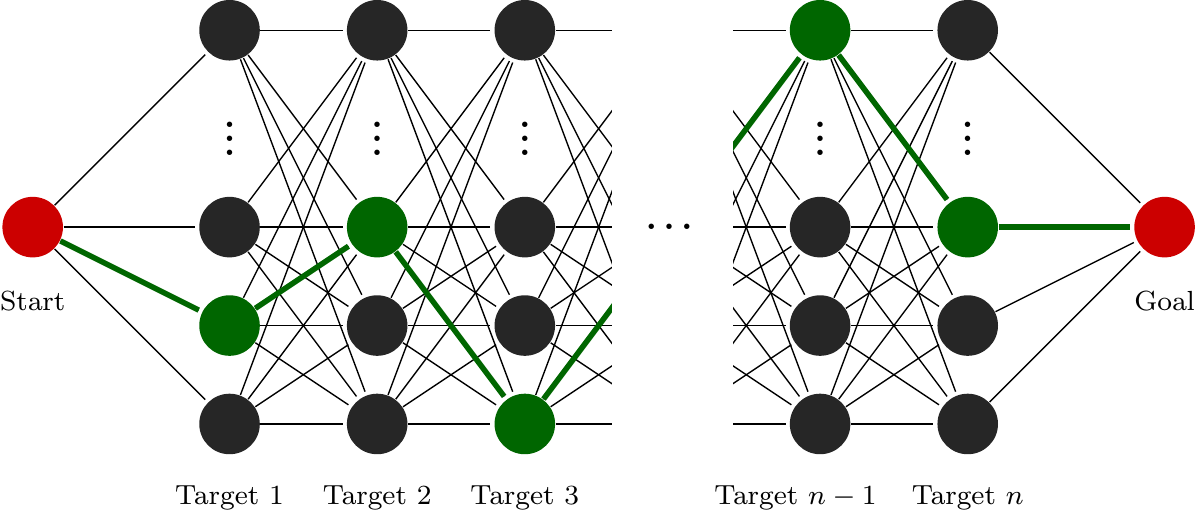}
  \caption{Graph constructed for Step~2 of the algorithm. The targets are
  ordered according to Step~1. The shortest path (green lines) connecting the
  \enquote{Start} and \enquote{Goal} vertices will yield the optimal
  configuration-space tour that visits the targets in the order specified by
  Step~1.}
  \label{fig:cspace_graph}
\end{figure}

Regarding Step~2, we first construct an undirected graph as depicted in
\fref{fig:cspace_graph}. Specifically, the graph has $n$ layers, each layer $i$
contains $m_i$ vertices representing the $m_i$ \ac{ik} solutions of target $i$
(the targets are ordered according to Step 1), resulting in a total of
$\sum_{i=1}^{n}{m_i}$ vertices. Next, for $i\in[1,\dots,n-1]$, we add an edge
between each vertex of layer $i$ and each vertex of layer $i+1$, resulting in a
total of $\sum_{i=1}^{n-1}{m_{i}m_{i+1}}$ edges. Finally, we add two special
vertices: \enquote{Start} and \enquote{Goal}, which are associated with the
robot \enquote{home} configuration, and connected respectively to the $m_1$
vertices of the first layer and the $m_n$ vertices of the last layer.

The edge costs are computed according to a configuration-space metric: for
instance, the cost for the edge joining vertices $\vb*{q}$ and $\vb*{q}'$
can be given by the Euclidean distance in the configuration space
$\sqrt{\sum_{k=1}^{\mdof}\qty(q_{k}-q'_{k})^2}$. \sref{sub:metrics_benchmark}
examines in detail how the choice of such metrics influences the quality of the
final path. One can note here that the metric should be fast to compute -- in
particular, collisions are ignored at this stage -- since the costs must be
computed for all $m_{1} + \sum_{i=1}^{n}{m_{i}m_{i+1}} + m_{n}$ graph edges.

Finally, we run a graph search algorithm to find the shortest path between the
\enquote{Start} and \enquote{Goal} vertices. By construction, any path between
the \enquote{Start} and \enquote{Goal} vertices will visit exactly one vertex in
each layer, in the order specified by Step 1. Conversely, for any choice of
\ac{ik} solutions for the $n$ targets, there will be a path in the graph between
the \enquote{Start} and \enquote{Goal} vertices and going through the vertices
representing these \ac{ik} solutions. Therefore, Step~2 will find the \emph{true
optimal} selection of \ac{ik} solutions that minimize the total cost, according
to the specified configuration-space metric, given the order of the targets.

\subsection{Complexity analysis}

For Step~1, it is well-known that \ac{tsp} is NP-complete, which means that
finding the true optimal tour for $n$ targets has in practice an exponential
complexity. Many heuristics have been developed over the years to find
near-optimal tours. For instance, \textit{2-Opt}~\cite{Croes1958} and
\ac{lkh}~\cite{Helsgaun2000} can find tours in practical times with an
optimality gap bellow $5\%$ and $1\%$ respectively \cite{Applegate2011}.

For Step~2, let $M$ be an upper-bound of the number of \ac{ik} solutions $m_i$
per target. The number of graph vertices is then $\order{nM}$ and the number of
the graph edges is $\order{nM^2}$. Since Dijkstra's algorithm (with binary heap)
has a complexity in $\order{\abs{E}\log\abs{V}}$ where $\abs{E}$ and $\abs{V}$
are respectively the number of edges and vertices, Step~2 has a complexity in
$\order{nM^2\log{\qty(nM)}}$.

For Step~3, one has to make $n-1$ queries to the motion planner, yielding a
complexity in $\order{n}$. However, as the constant in the $\order{}$ (average
computation time per motion planning query) is large, the overall computation
time is dominated by that of Step~3 in our setting. In general, the computation
time of motion planning queries depends largely on the environment (obstacles),
see~\cite{Sucan2012,Meijer2017} for recent benchmarking results showing the CPU
time required when planning practical robot motions.


\section{Experiments}
\label{sec:experiments}

This section evaluates the proposed method when applied to the drilling task
shown in \fref{fig:setup}. Our system is formed by a Denso VS060 6-\ac{dof}
industrial manipulator equipped with a commercial off-the-shelf hand drill. All
benchmarks were executed in a system with Intel\textsuperscript{\textregistered}
Core\texttrademark~i7 processor and 24 GB RAM, GeForce GTX 960M video card,
running Ubuntu 16.04 (Xenial), 64 bits.

\subsection{Benchmarking task-space \ac{tsp} solvers}
\label{sec:tsp_comparison}

To solve the task-space \ac{tsp} (Step~1 of our algorithm), one may use exact or
near-optimal solvers. The choice depends on the trade-off between the available
CPU time and the solution quality. Here we evaluate three \ac{tsp} solvers.

\begin{enumerate}

  \item \textit{Exact}: \ac{cip} can be used to find true optimal \ac{tsp}
  tour~\cite{Applegate2011}. Here, we used the \acs{scip} Optimization Suite
  \cite{Achterberg2009} to implement an exact solver;

  \item \textit{2-Opt}: We re-implemented this simple, yet efficient, algorithm
  to find a near-optimal solution to \ac{tsp}. The algorithm iteratively
  improves an initial guess by repeatedly replacing pairs of edges that cross
  over \cite{Applegate2011}.

  \item \textit{RNN}: We re-implemented this algorithm, which consists in
  iteratively selecting the nearest neighbor as the next target to visit. This
  process is repeated until all the targets are visited \cite{Applegate2011}.
  One can do several \emph{restarts} from different initial targets, and choose
  the tour with the lowest cost from all the restarts. The drawback of this
  method is that it tends to corner itself, which requires long edges to get
  back to unvisited targets.

\end{enumerate}

\begin{figure}[t]
  \centering
  \vspace*{2mm}
  \subfloat[]{\includegraphics[height=40mm]{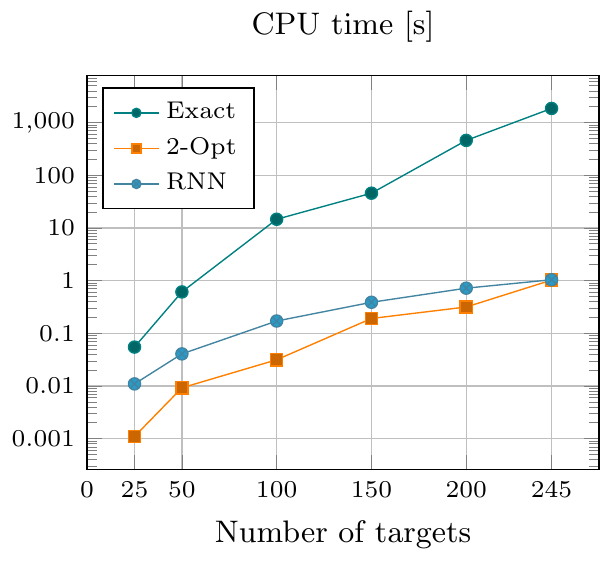}}\;
  \subfloat[]{\includegraphics[height=40mm]{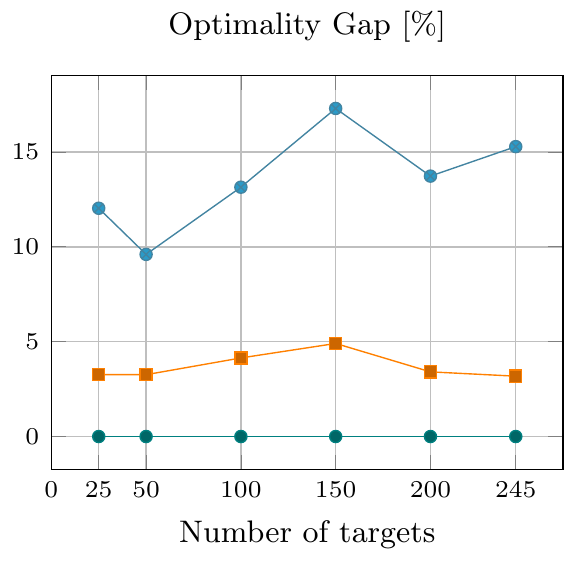}}
  \caption{Benchmarking results for three task-space \ac{tsp} solvers. (a) CPU
  times for different number of targets. The Y axis is logarithmic (b) Task
  execution time for different number of targets. The sets of targets are
  selected randomly. The last sample considers all the $245$ targets.}
  \label{fig:tsp_solvers}
\end{figure}

\fref{fig:tsp_solvers} shows a benchmark of the three methods. We run the
\ac{tsp} solvers on task-space subsets of $25$, $50$, $100$, $150$, $200$ random
targets as well as on the total $245$ targets.  One can observe that the
\textit{2-Opt} solver yields high-quality tours (less than $5\%$ of
sub-optimality) with low CPU time usage (less than $1$\,s). As for the
\textit{Exact} solver, it is not practical for more than $150$ targets.
Therefore, for all the subsequent experiments, we shall use \textit{2-Opt} as
our near-optimal task-space \ac{tsp} solver.

\subsection{Benchmarking configuration-space metrics}
\label{sub:metrics_benchmark}

The \textit{configuration-space metric} that defines the edge cost in Step~2 of
our algorithm is the key component for the overall performance of the method.
Given two robot configurations $\vb*{q}$ and $\vb*{q}'$, the ideal cost of the
edge $c^{*}\qty(\vb*{q}, \vb*{q}')$ is the duration of a time-optimal
collision-free trajectory between them. However, since there are thousands of
such edges, running full-fledged motion planning algorithms (with collision
checks) for every edge would not be tractable. Therefore, one must consider
approximate metrics, which should be fast to compute, yet give a good prediction
of the corresponding time-optimal collision-free trajectory duration. Here we
evaluate three such metrics.

\begin{figure}[t]
  \centering
  \vspace*{2mm}
  \subfloat[]{\includegraphics[height=40mm]{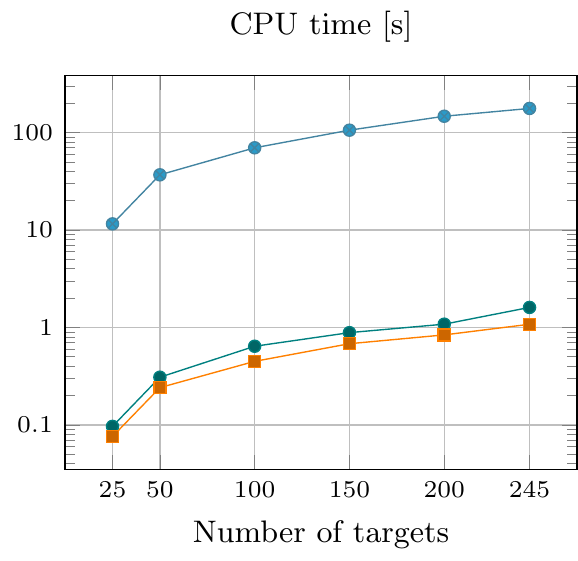}}\;
  \subfloat[]{\includegraphics[height=40mm]{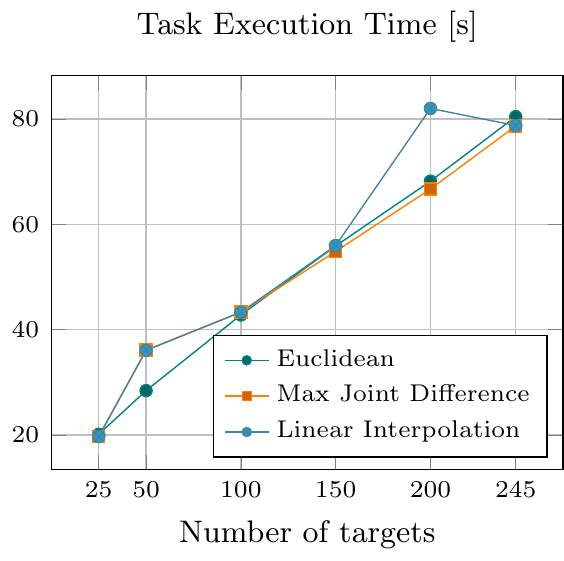}}
  \caption{Benchmarking results for three C-space metrics. (a) CPU times for
  different number of targets. The Y axis is logarithmic (b) Task execution time
  for different number of targets. The sets of targets are selected randomly.
  The last sample considers all the $245$ targets.}
  \label{fig:metrics_benchmark}
\end{figure}

\begin{enumerate}

  \item \textit{Weighted Euclidean joint distance}: The cost $c\qty(\vb*{q},
  \vb*{q}')$ is estimated as the weighted $\mbox{L}^{2}$ norm:
  \begin{equation*}
    c\qty(\vb*{q}, \vb*{q}') :=
    \sqrt{\sum_{k=1}^{\mdof}w_{k}\qty(q'_{k}-q_{k})^2}\; ,
  \end{equation*}
  where $w_{k}$ is a positive weight for joint $k$. The weights are chosen in
  proportion to the maximum possible distance (Euclidean distance in the
  task-space) traveled by any point on the robot, when moving along the
  corresponding joint. Similar to \cite{Bohlin2000}, in our experiments this
  metric outperforms consistently the Euclidean joint distance.

  \item \textit{Maximum joint difference}: The cost $c\qty(\vb*{q}, \vb*{q}')$
  is estimated as follows:
  \begin{equation*}
    c\qty(\vb*{q}, \vb*{q}') :=
          \max_{k}\abs{\frac{\qty(q'_{k}-q_{k})}{\dot{q}^{\max}_k}}\; .
  \end{equation*}
  The intuition of this metric is to determine the maximum joint displacement
  when \enquote{moving} from $\vb*{q}$ to $\vb*{q}'$ by simply computing the
  joint difference, $\qty(q'_{k}-q_{k})$, over the joint $k$ velocity limit, for
  $k \in \qty[1,\dots,\mdof]$. Then the maximum value is used;

  \item \textit{Linear trajectory interpolation}: the cost $c\qty(\vb*{q},
  \vb*{q}')$ is given by the duration of a trajectory obtained by linear
  interpolation. It only requires to specify the positions, $\vb*{q}$ and
  $\vb*{q}'$, and guarantees continuity at the position level subject to the
  robot velocity and acceleration bounds. Moreover, this metric does not
  consider obstacles which greatly reduces its computing time.

\end{enumerate}

\fref{fig:metrics_benchmark} shows the benchmarking results for the
three proposed configuration-space metrics. One can observe that the
\textit{Maximum joint difference} metric takes the lowest CPU time and
yields task execution times comparable to, in some cases even better
than, the \textit{Euclidean} and \textit{Linear Interpolation
  metrics}. Therefore, for all the subsequent experiments, we shall
use \textit{Maximum joint difference} as our metric.

\subsection{Benchmarking discretization step size for the free \ac{dof}}
\label{sub:discrete_benchmark}

Many industrial tasks such as spot-welding, spay-painting or drilling involve
less than 6 degrees of freedom. Therefore, a classical 6-\ac{dof} industrial
robot has more joints than strictly required to execute such tasks.
Specifically, the drilling task at hand involves 5 \ac{dof}, since the rotation
$\theta$ about the drilling direction is irrelevant. One approach to tackle this
redundancy can consist in setting a specific value for the irrelevant \ac{dof}:
for instance $\theta \in \{0, \frac{\pi}{2}, \pi, \frac{3\pi}{2}\}$ for a
$\frac{\pi}{2}$ discretization step size. For each of the discretized value of
$\theta$, we then have a full 6-\ac{dof} \ac{ik} problem. To solve the full
6-\ac{dof} \ac{ik}, we next use OpenRAVE's IKFast \cite{diankov2010}, which
outputs all the \ac{ik} solutions (here we have a \enquote{discrete} redundancy
situation -- think of the \enquote{elbow up} and \enquote{elbow down}
configurations). We finally group all the IK solutions corresponding to all the
discretized values of $\theta$ into a single list, which is the list of all IK
solutions that will be considered for a given target. \tref{tab:discret_effect}
gives the total number of IK solutions considered as a function of the
discretization step size and of the number of targets.

\renewcommand{\arraystretch}{1.5}
\begin{table}[htp]
\caption{Number of robot configurations depending on the discretization
step size and the number of targets}
\label{tab:discret_effect}
\centering
\begin{tabular}{c c c c c c c}
\toprule
\textbf{Step} & \multicolumn{6}{c}{\textbf{Number of targets}}  \\
\textbf{Size} & \bm{$25$} & \bm{$50$}  & \bm{$100$} & \bm{$150$} & \bm{$200$} & \bm{$245$}\\
\midrule
$\pi$            & $235$   & $538$   & $1,044$ & $1,548$  & $2,071$  & $2,495$\\
$\frac{\pi}{2}$  & $344$   & $816$   & $1,598$ & $2,368$  & $3,094$  & $3,779$\\
$\frac{\pi}{3}$  & $481$   & $1170$ & $2,287$ & $3,362$  & $4,424$  & $5,418$\\
\bm{$\frac{\pi}{4}$} & \bm{$623$} & \bm{$1,505$} & \bm{$2,939$} & \bm{$4,377$} & \bm{$5,668$} & \bm{$6,990$}\\
$\frac{\pi}{6}$  & $952$   & $2,236$ & $4,417$ & $6,528$  & $8,471$  & $10,421$\\
$\frac{\pi}{12}$ & $1,934$ & $4,428$ & $8,724$ & $12,981$ & $16,811$ & $20,720$\\
\bottomrule
\end{tabular}
\end{table}

\begin{figure}[t]
  \centering
  \vspace*{2mm}
  \subfloat[]{\includegraphics[height=40mm]{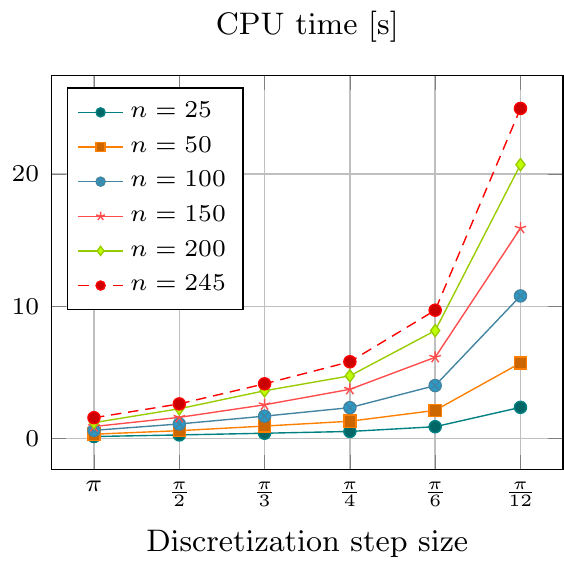}}\;
  \subfloat[]{\includegraphics[height=40mm]{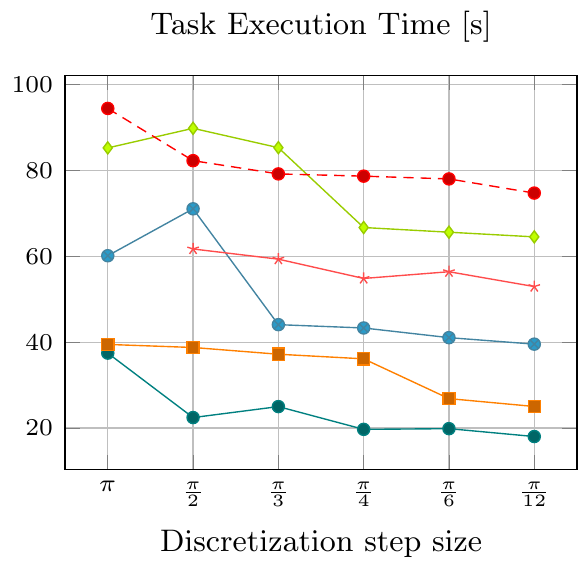}}
  \caption{Benchmarking results showing the effect of the discretization step size
  of the free \ac{dof}.}
  \label{fig:discrete}
\end{figure}

One can see that the choice of the \textit{discretization step size} is governed
by a trade-off between speed and optimality. \fref{fig:discrete} shows the
computation time and task execution time as a function of the discretization
step size. As expected, the computation time increases as the discretization
step size decreases, but interestingly, the task execution time does not change
significantly for step sizes below $\frac{\pi}{4}$, which thus yields a good
trade-off between CPU time and task execution time. Therefore, for all the
subsequent experiments, we shall use $\frac{\pi}{4}$ as our discretization step
size.

\subsection{Comparison to other methods}
\label{sub:other_methods}

\begin{figure}[t]
  \centering
  \subfloat{\includegraphics[height=40mm]{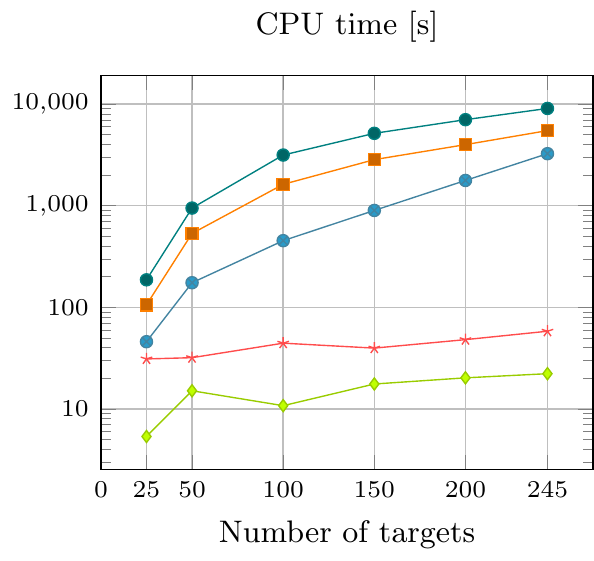}}\;
  \subfloat{\includegraphics[height=40mm]{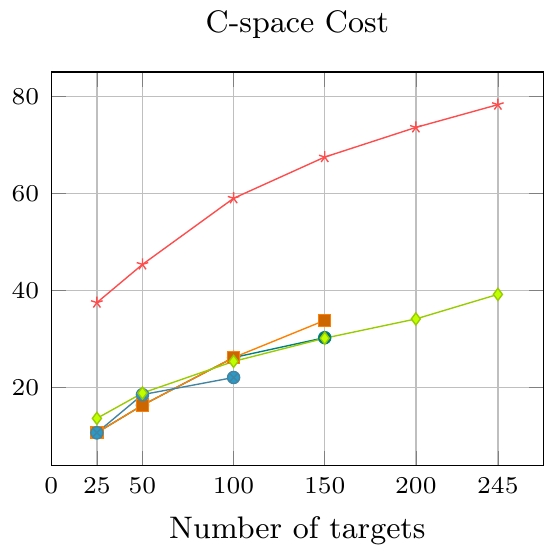}}  \\
  \subfloat{\includegraphics[height=40mm]{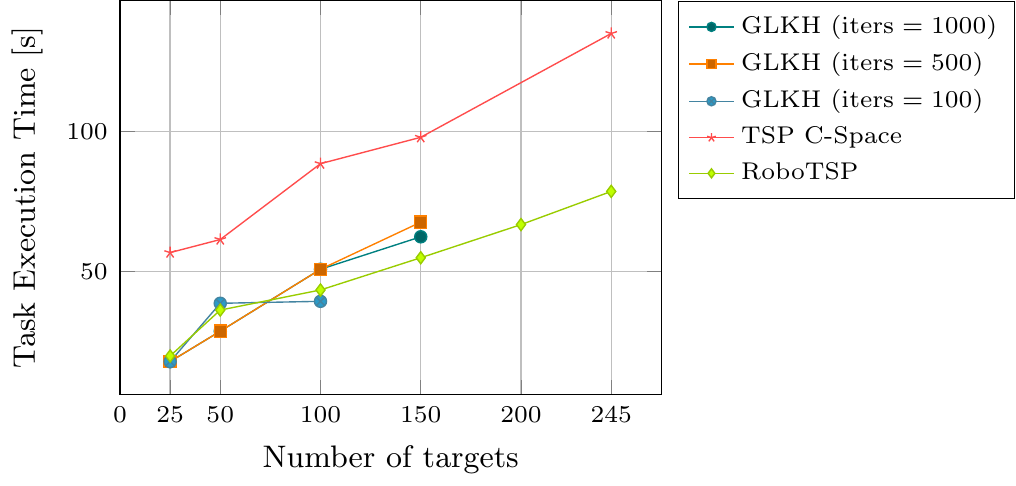}}
  \caption{Comparison of our method (\textit{RoboTSP}) against other methods: i)
  Solved as \acs{tsp} in configuration-space; ii) Solved as \acs{gtsp} using
  \acs{glkh}.}
  \label{fig:other_methods}
\end{figure}

As none of the methods described in \sref{sec:related} provide public
implementations, we were unable to reproduce their results and perform a fair
comparison with the method herein presented. However, we can note that the
computation times reported in previous works are several orders of magnitude
higher than ours, yet for problem instances that are smaller\,\footnote{The
problem instance size is considered in terms of targets and configurations per
target} than what we consider in this work.

In the following, we compare our method to two of the existing methods, which we
could re-implement.

\begin{enumerate}
  \item \textit{\acs{tsp} in C-space}~\cite{Edan1991}: when only one
  configuration is considered per target, \ac{rtsp} is reduced to a regular
  \ac{tsp} in the configuration space. Here, for each target, we consider the
  \ac{ik} solution with the best manipulability index~\cite{Yoshikawa1985};
  \item \textit{\acs{glkh}}: here the \ac{rtsp} is formulated as a
  \ac{gtsp}~\cite{Saha2006,Wurll1999,Wurll2001}. To solve that \ac{gtsp}, we use
  the state-of-the-art \acs{glkh} solver~\cite{Helsgaun2015} which makes use of
  the \ac{lkh} heuristic~\cite{Helsgaun2000}.
\end{enumerate}

\fref{fig:other_methods} shows the comparison of our method (\textit{RoboTSP})
to the two methods just described. While the \textit{\acs{tsp} in C-space}
method has a similar running time as \textit{RoboTSP} (indeed both run a
\ac{tsp} on the same number of targets), the time durations of the trajectories
it produces are higher than those of \textit{RoboTSP}, since it does not
optimize the \ac{ik} choice per target.

\textit{\acs{glkh}} produces trajectories with similar total duration as
\textit{RoboTSP} but the computing time is higher several orders of magnitude.

A visualization with the trajectories produced by these three methods to visit
all the $245$ targets is available at \url{https://youtu.be/w33QfRjKFs8}.


\section{Conclusions}
\label{sec:conclusions}

We have proposed a method to determine a near-optimal sequence to
visit $n$ targets with multiple configurations per target, also known
as the \acl{rtsp}. For a complex drilling task, which requires
visiting 245 targets with an average of 28.5 configurations per
target, our method could compute a high-quality solution in less than
a minute. To our knowledge, no existing approach could have solved the
same problem in practical times. We have also provided a carefully
benchmarked open-source software solution, which can be readily used
in complex, real-world, applications such as drilling, spot-welding or
spray-painting.


\section*{Acknowledgment}
This work was supported in part by NTUitive Gap Fund NGF-2016-01-028
and SMART Innovation Grant NG000074-ENG.

\IEEEtriggeratref{6}
\bibliographystyle{IEEEtran}
\bibliography{IEEEabrv,references}

\end{document}